\pdfoutput=1

\documentclass[11pt]{article}

\usepackage[final]{coling}

\usepackage{times}
\usepackage{latexsym}
\usepackage{array}
\usepackage{graphicx}
\usepackage{epstopdf}
\usepackage{makecell}
\newcolumntype{L}[1]{>{\raggedright\let\newline\\\arraybackslash\hspace{0pt}}m{#1}}
\newcolumntype{C}[1]{>{\centering\let\newline\\\arraybackslash\hspace{0pt}}m{#1}}
\newcolumntype{R}[1]{>{\raggedleft\let\newline\\\arraybackslash\hspace{0pt}}m{#1}}
\usepackage{multirow}
\usepackage{svg}
\usepackage{makecell}
\usepackage[utf8]{inputenc}
\usepackage{amsmath}
\usepackage{cleveref}
\crefname{section}{§}{§§}
\Crefname{section}{§}{§§}
\usepackage{booktabs}
\usepackage{xcolor} 
\usepackage{colortbl}
\usepackage{color}
\usepackage{algorithm}
\usepackage{algorithmic}
\usepackage{pifont}
\usepackage{arydshln}
\usepackage{lipsum}
\usepackage{tcolorbox}

\usepackage[T1]{fontenc}

\usepackage[utf8]{inputenc}

\usepackage{microtype}

\usepackage{inconsolata}

\usepackage{graphicx}

\title{\textit{SKIntern}: Internalizing Symbolic Knowledge for Distilling Better CoT Capabilities into Small Language Models}

\author{Huanxuan Liao$^{1,2}$, Shizhu He$^{1,2}$\thanks{Corresponding author}, Yupu Hao$^{1,2}$, Xiang Li$^{1,2}$, \\ \textbf{Yuanzhe Zhang}$^{4}$, \textbf{Jun Zhao}$^{1,2}$, \textbf{Kang Liu}$^{1,2,3}$  \\
    $^1$ The Key Laboratory of Cognition and Decision Intelligence for Complex Systems, \\
    Institute of Automation, Chinese Academy of Sciences, Beijing, China \\
    $^2$ School of Artificial Intelligence, University of Chinese Academy of Sciences, Beijing, China \\
    $^3$ Shanghai Artificial Intelligence Laboratory, Shanghai, China \\
    $^4$ National Science Library, Chinese Academy of Sciences, Beijing, China \\
  {\{liaohuanxuan2023, haoyupu2023, lixiang2022\}@ia.ac.cn} {\{shizhu.he, kliu, jzhao\}@nlpr.ia.ac.cn} \\}

\begin{document}
\maketitle
\begin{abstract}
Small Language Models (SLMs) are attracting attention due to the high computational demands and privacy concerns of Large Language Models (LLMs).
Some studies fine-tune SLMs using Chains of Thought (CoT) data distilled from LLMs, aiming to enhance their reasoning ability.
Furthermore, Some CoT distillation methods introduce external symbolic knowledge into the generation process to improve the limited knowledge memory, reasoning ability and out-of-domain (OOD) generalization of SLMs. 
However, the introduction of symbolic knowledge increases computational overhead and introduces potential noise.
In this paper, we introduce \textit{SKIntern}, an innovative approach that empowers SLMs to internalize symbolic knowledge and few-shot examples gradually through a progressive fine-tuning process, guided by a predefined linear decay schedule under curriculum learning. 
By efficiently internalizing knowledge, \textit{SKIntern} reduces computational overhead and speeds up the reasoning process by focusing solely on the question during inference. It outperforms state-of-the-art baselines by over 5\%, while reducing inference costs (measured in FLOPs) by up to $4\times$ across a wide range of SLMs in both in-domain (ID) and out-of-domain (OOD) tasks.
Our code will be available at \url{https://github.com/Xnhyacinth/SKIntern}.
\end{abstract}

\section{Introduction}

Large Language Models (LLMs) \cite{llama, Qwen2TR} have greatly excelled at various complex reasoning tasks such as mathematical \cite{gsmplus}, symbolic \cite{bbh} and logical \cite{InvestigatingSCSymbolic} reasoning, by applying Chains of Thought (CoT) prompting \cite{chain} and In-Context Learning (ICL) \cite{iclcompositional, automatic}. Nonetheless, the high computational expenses and data privacy issues associated with LLMs have highlighted the need for Small Language Models (SLMs) \cite{dissurvey}. However, these advanced reasoning and knowledge capabilities are typically modeled in larger models ($\geq$13B), making it challenging to replicate in SLMs ($\leq$7B) \cite{scaling}. 

\begin{figure}[t]
\centerline{\includegraphics[width=0.5\textwidth]{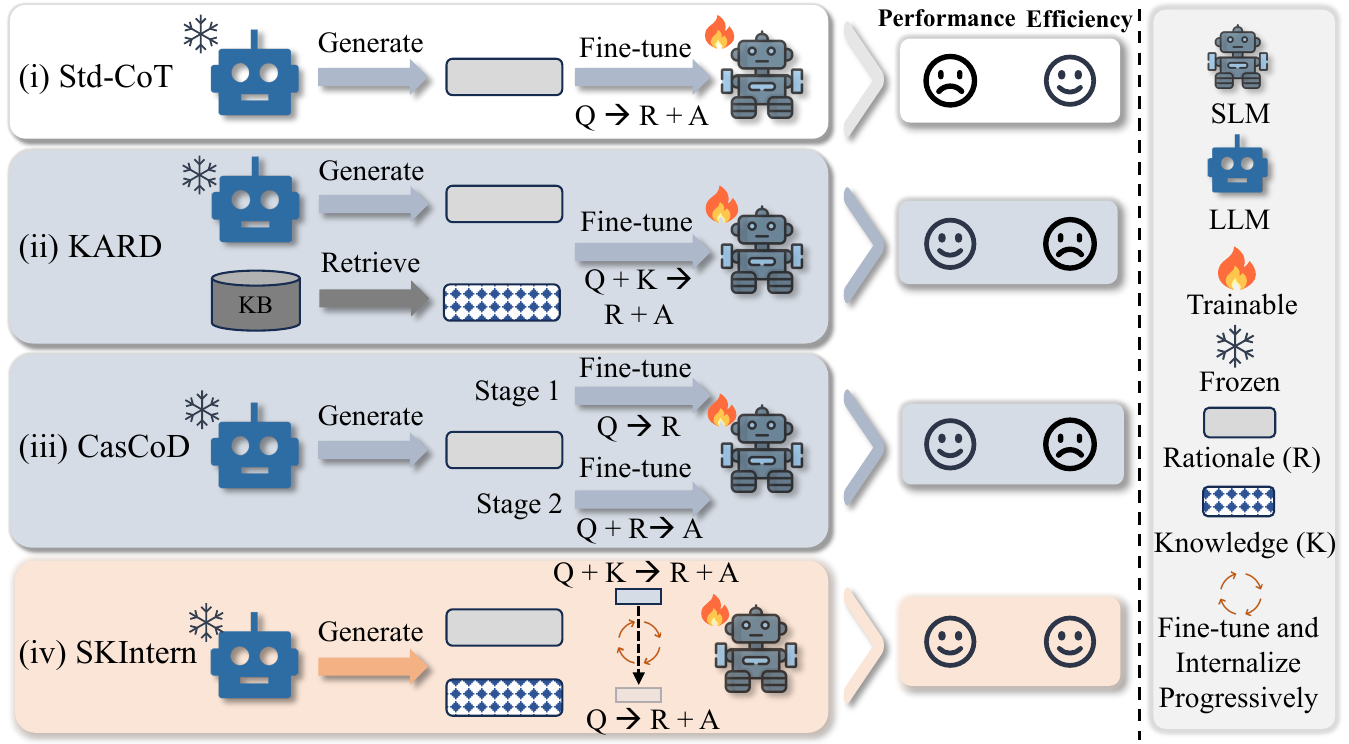}}
\caption{Knowledge utilization comparisons of  \textit{SKIntern} and other typical CoT distillation methods.  (i) Std-CoT: SLM is fine-tuned to generate the rationale and answer for the question (Q -> R + A). (ii) KARD: Fine-tune the SLM to generate the rationale and answer based on the question and the retrieved symbolic knowledge (Q + K -> R + A). (iii): CasCoD: Decompose the single CoT learning step into two comprehensive learning steps of rationale generation (Q -> R) and rationale utilization (Q + R -> A). (iv): \textit{SKIntern}: Like human interns, SLMs gradually absorb and internalize symbolic knowledge provided by LLMs during the progressive fine-tuning, thereby achieving efficient (Q -> R + A) and effective reasoning (\textit{modeling K in parameters}).
}
\label{intro}
\end{figure}

To improve the reasoning ability of SLMs, existing works \cite{Fu2023SpecializingSL, mt-cot} aim to distill the reasoning ability of LLMs into SLMs by fine-tuning SLMs with high-quality rationales obtained from LLMs, known as standard CoTs distillation (Std-CoT) \cite{stdcot}. However, due to the limited parameter size of SLMs, they cannot effectively memorize all knowledge and model reasoning ability, making it difficult to generalize to out-of-domain (OOD) tasks.

Recently, several methods have been proposed to further improve the knowledge memory and reasoning ability of SLMs. For example, as illustrated in Figure \ref{intro}, KARD \cite{kard} uses external knowledge bases to enhance the memory capacity of SLMs, while CasCoD \cite{improve} employs cascading decomposition to support gradual learning. However, those methods lead to two challenges: 1) \textbf{Redundant and noisy symbolic knowledge degrades the effect of CoT distillation}. Document retrieval based on similarity frequently results in repetitive and trivial content, complicating the model's ability to extract key information \cite{LostIT}. Additionally, retrieved documents often contain irrelevant or misleading information, introducing noise that diminishes the model's performance. 2) \textbf{Long input and multi-stage generation reduce the inference efficiency of CoT distillation}. Processing additional documents and rationales imposes significant memory and computational burdens, and the complex inference process complicates deployment and implementation, reducing overall efficiency.
Therefore, a key challenge of CoT distillation is: \textit{\textbf{Can we effectively and efficiently transfer the rich knowledge and reasoning ability of LLMs through CoT distillation while minimizing computational overhead?}}

To resolve the above challenge, we examine the human learning process and draw analogies to model fine-tuning.
For instance, at first, an intern typically needs detailed explanations, examples, and documentation to learn new skills~\cite{promptintern}. However, once they have internalized this knowledge and mastered the required skills, such extensive information is no longer needed. 
Therefore, we believe that if SLMs are provided with detailed guidance and symbolic knowledge while learning rationales from LLMs, their learning outcomes can be greatly enhanced. By gradually internalizing this knowledge into their parameters, SLMs can independently develop efficient reasoning abilities, eliminating the need for additional document retrieval or multi-stage generation.

To perform an efficient and effective CoT distillation, we introduce a novel approach \textbf{\textit{SKIntern}} that internalizes the symbolic knowledge during model fine-tuning and enables efficient inference without additional context. Specifically, our method comprises two key steps. Initially, for each training instance, LLMs generate rationales and symbolic knowledge (such as the learning summaries and supplementary materials) and we select the most relevant ones using cosine similarity.
Secondly, we gradually perform token-level symbolic knowledge compression and instance-level example pruning based on a predefined linear decay schedule. This refined information is then used to fine-tune the SLM to generate the rationale from the LLMs and the answer.
As the schedule progresses, both symbolic knowledge and examples are internalized into the model's parameters, enabling effective reasoning based solely on the questions during inference.

We evaluate \textit{SKIntern} on open-source models like TinyLLaMA \cite{TinyLlamaAO} and LLaMA2-7B \cite{llama} across factual, mathematical, and general reasoning benchmarks. By internalizing symbolic knowledge into parameters and addressing questions exclusively during inference, \textit{SKIntern} surpasses strong baselines in both ID and OOD tasks while significantly reducing computational requirements (measured in FLOPs).
This supports our hypothesis that internalizing symbolic knowledge can significantly reduce inference costs, thereby avoiding explicit processing during inference. Additionally, we find that the performance of \textit{SKIntern} can be further enhanced by incorporating few-shot examples into parameters with minimal additional computation. These improvements suggest that our method balances efficiency and effectiveness, making it highly suitable for optimizing SLM inference performance in cost-sensitive scenarios.
In conclusion, the contributions of this paper are summarized as follows:
\begin{itemize}
    \item We propose a novel CoT distillation method \textit{SKIntern} designed to emulate the incremental learning process of interns, gradually learning and mastering knowledge and skills.
    \item We progressively internalize the symbolic knowledge generated by the LLM and the selected examples into parameters, thereby achieving effective and efficient inference without the need for additional information.
    \item We conducted extensive experiments on 7 reasoning benchmarks. \textit{SKIntern} outperforms robust baselines by $5\%$ in both ID and OOD tasks, while reducing inference costs by up to $4\times$ across a broad spectrum of SLMs.
\end{itemize}

\section{Related Work}

\noindent \textbf{CoT Distillation} transfers the reasoning ability of LLMs to SLMs, where reasoning ability is an emergent property that enables LLMs to excel in reasoning tasks through Chains of Thought (CoT) prompting (e.g., Let’s think step-by-step) \cite{chain, LargeLM}. Recent works \cite{stdcot, Fu2023SpecializingSL} show that this CoT inference mechanism can be used for distillation: fine-tuning a smaller student model using CoT sequences extracted from a larger teacher model significantly boosts performance. Further studies \cite{distilling, mt-cot} have proposed treating the learning of rationales and answers as distinct optimization objectives. However, these approaches often overlook the limited memory and reasoning ability of SLMs, making it difficult to generalize to OOD tasks.
KARD \cite{kard} boosts SLMs' memory by retrieving external knowledge, while CasCoD \cite{improve} refines rationale perception through cascading decomposition learning. However, both methods require processing more tokens (document retrieval and multi-stage generation), which introduces additional complexity and uncontrollability in reasoning tasks. Our proposed method mirrors how interns learn a new task by first providing full symbolic knowledge and examples and gradually internalizing them into the parameters, achieving effective inference without additional information. 

\noindent \textbf{Prompt Compression} condenses lengthy prompts, retaining only essential information while reducing length. This process can be divided into three main methods: Information entropy-based techniques \cite{compressing, longllmlingua} 
use a small language model to calculate the self-information or error-proneness of tokens, removing those with lower error-proneness; Soft prompts methods \cite{adapting, Mu2023LearningTC} require fine-tuning LLM parameters to use learnable tokens for condensing prompts; Interpretable summaries methods \cite{RECOMPIR, LLMLingua2DD} extract data from the LLM to train models for generating more interpretable text summaries. A method analogous to ours is PromptIntern \cite{promptintern}, which achieves prompt compression through progressive fine-tuning. We internalize knowledge and examples into the parameters by gradually pruning the prompt during training, allowing the prompt to be discarded during inference.

\section{Methodology}

\begin{figure*}[t]
\centerline{\includegraphics[width=1.0\textwidth]{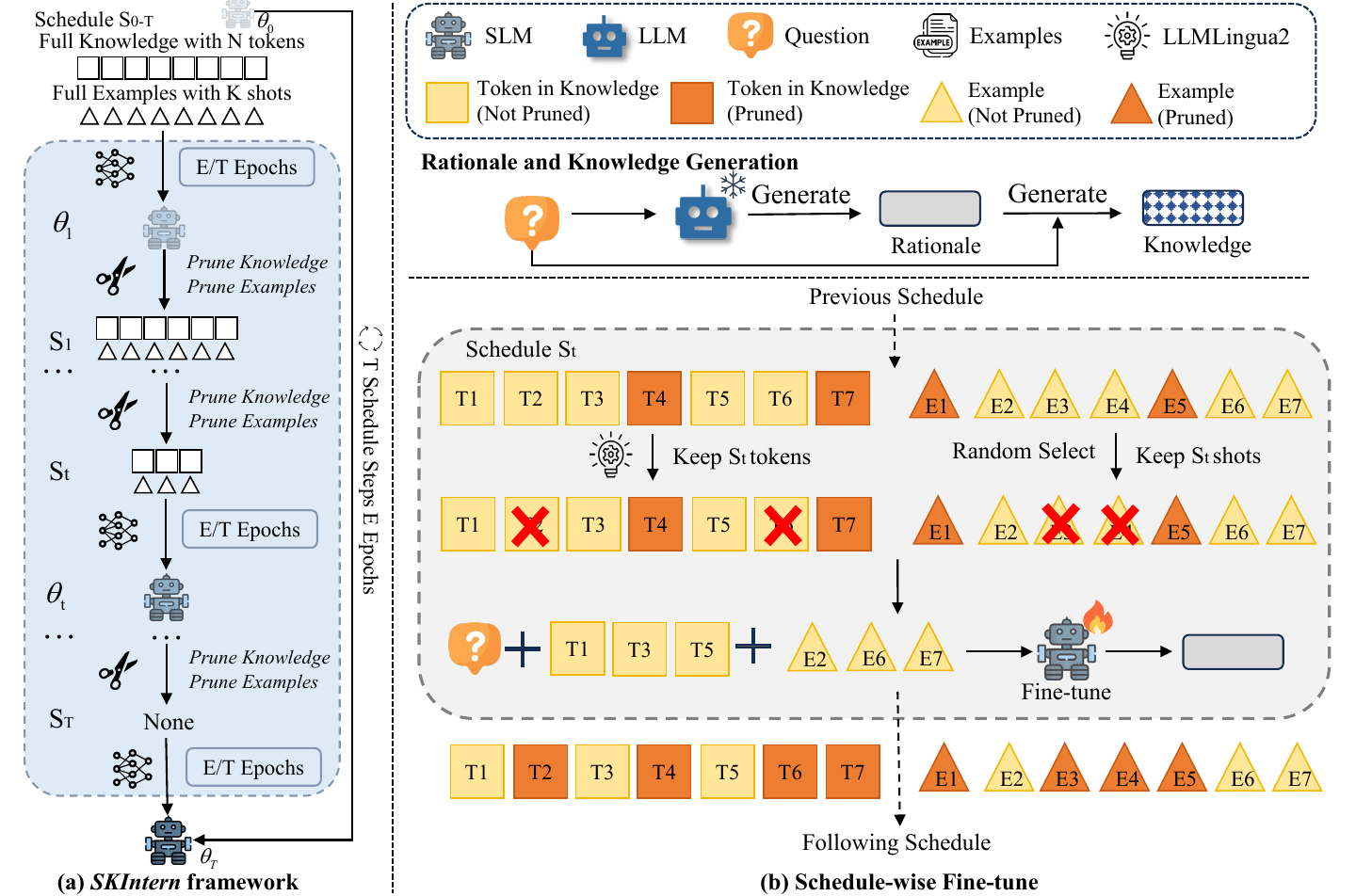}}
\caption{ Overview of the \textit{SKIntern} framework. \textit{SKIntern} starts with full symbolic knowledge and examples, and progressively prunes them to gradually internalize knowledge, reducing the prompt length and the number of computations towards the SLM. Based on schedule $\mathcal{S}$, we perform effective knowledge compression and example pruning before fine-tuning the SLM to generate rationales and answers. Gradual fine-tuning makes SLMs internalize knowledge and examples into parameters, thereby enhancing performance without increasing computational cost.
}
\label{model}
\end{figure*}

In this section, we introduce the detailed procedures of \textit{SKIntern}. As illustrated in Figure \ref{model}, \textit{SKIntern} starts with the full knowledge and examples, and progressively prunes tokens to gradually internalize them into the model's parameters, reducing the prompt length and the number of computations towards the model. Below, we first describe how to extract CoT and symbolic knowledge from the teacher LLM in \cref{extrat}. Then we introduce techniques for symbolic knowledge compression and examples pruning to convert them into parameters in \cref{progress}. Finally, we present a customized progressive fine-tuning pipeline for \textit{SKIntern} in \cref{pipeline}. Note,
\textit{SKIntern} achieves great results without additional knowledge and examples in input compared with Std-Cot during inference, merely depending on the knowledge stored in the parameters.

\subsection{Rationale and Knowledge Generation}
\label{extrat}

\noindent \textbf{Rationale Generation.} In our problem setup, we assume a given training dataset $\mathcal{D}_{\text{train}} = \left\{ (\boldsymbol{x}_i, \boldsymbol{y}_i) \right\}_{i=1}^n$ for the target task, where $\boldsymbol{x}_i$ is the input sequence (question in QA) and $\boldsymbol{y}_i$ is the label (answer in QA). LLMs can generate high-quality rationales, which is known as the emergent ability \cite{LargeLM}. Our objective is to transfer this capability to SLMs through CoT distillation. Firstly, we leverage the CoT prompting \cite{chain} to guide the teacher LLM in generating proper $\boldsymbol{l}$ rationales for each training data point: $\boldsymbol{r}_{ij} = \text{LLM}(\boldsymbol{p}_c, \boldsymbol{x}_i, \boldsymbol{y}_i)$ where $\boldsymbol{r}$ are generated rationales, $j \in \{1,...,\boldsymbol{l}\}$ and $\boldsymbol{p}_c$ is the prompt which is shown in Appendix \ref{prompt_cot}. 
To maintain high-quality CoT data, we filter out reasoning processes that do not yield correct results, retaining only the distilled CoT sequences that lead to accurate outcomes as the training data \cite{distilling}.

\noindent \textbf{Symbolic Knowledge Generation.} Rationales offer insights into the logic behind answers, which is crucial for SLMs to respond more precisely. However, SLMs with limited parameters may struggle to retain all training data and complex reasoning capabilities, which can affect the quality of rationale generation \cite{kard}. Furthermore, this single learning might lead SLMs to focus on directly answering questions after reading, potentially impairing their ability to generalize in reasoning \cite{improve}. Hence, it is imperative to present the SLM with knowledge in the initial stages of learning to facilitate its understanding. 

We use prompt $\boldsymbol{p}_k$ which is in the Appendix \ref{prompt_know} to enable teacher LLM to generate learning summaries ${\boldsymbol{k}^m}$ that incorporate thinking processes and supplemental knowledge ${\boldsymbol{k}^p}$, collectively referred to as symbolic knowledge $\boldsymbol{k}$. Formally, the teacher LLM generate $\boldsymbol{m}$ knowledge using the question $\boldsymbol{x}_i$, the rationale $\boldsymbol{r}_i$ and the answer $\boldsymbol{y}_i$: $\boldsymbol{k}_{ij} = \text{LLM}(\boldsymbol{p}_k, \boldsymbol{x}_i, \boldsymbol{y}_i, \boldsymbol{r}_i)$, where $j \in \{1,...,\boldsymbol{m}\}$. A rationale typically addresses a specific question, whereas knowledge generally offers broader explanations, methods and outlines.

\subsection{Progressive Internalization} 
\label{progress}

Before this work, knowledge augmentation has been successfully applied to optimize SLM inference \cite{kard}. However, these methods necessitate processing full knowledge during both training and inference phases, significantly increasing computation overhead. Consequently, they are unsuitable for scenarios with limited computational resources. In contrast, 
by pruning the number of tokens gradually during the training phase, SKIntern processes only the question during inference without requiring additional symbolic knowledge.

We implement a predefined schedule $\mathcal{S}$ to regulate the pruning rate of knowledge and examples. At each step, the pruned symbolic knowledge and few-shot examples are appended to the question, fine-tuning the SLM over $E/T$ epochs, where the total training spans $E$ epochs. As shown in Figure \ref{model} (a), with $T$ total schedule steps, the value of $\mathcal{S}$ progressively decreases from 1 to 0. As the compression rate increases and fine-tuning progresses, the knowledge in the input gradually reduces to 0, leading to the internalization of knowledge into the model's parameters.


\noindent \textbf{Symbolic Knowledge Compression.} Inspired by prompt compression works \cite{LLMLingua2DD}, we aim to gradually increase the compression rate to reduce the symbolic knowledge at the token-level determined by $\mathcal{S}_t$ at $t$-th step and internalize it into the parameters, which can be expressed as:
\begin{equation}
    \boldsymbol{k}_i^t = \text{LLMLingua2}(\boldsymbol{k}_i, \mathcal{S}_t)
\end{equation}
where LLMLingua2\footnote{We apply LLMLingua-2 as the default compressor as it performs the best before June 2024.} \cite{LLMLingua2DD} is a task-agnostic prompt compression method that distills knowledge from the LLM and fine-tunes the encoder to compress prompts without losing key information, 
$\boldsymbol{k}^t$ is the compressed symbolic knowledge at $t$-th step, varying at different schedule $\mathcal{S}_t$. Considering that prompt compression is our means of pruning, we directly utilize existing prompt compression methods and models to achieve the compression of knowledge at different learning schedule steps.

\noindent \textbf{Example Pruning.} During inference, incorporating few-shot examples can significantly enhance model performance, and incorporating these examples into the fine-tuning stage can further improve the comprehension of various task inputs and outputs \cite{promptintern}. However, directly adding verbose minority examples to the input would increase the load on the context window and elevate inference computation and latency. So we propose a similarity-based instance-level pruning method to internalize the examples into parameters.
For each training instance $(\boldsymbol{x}_i, \boldsymbol{y}_i)$, we begin by employing a relevance scoring function $sim(\cdot, \cdot)$ to assess the similarity between its and different instances in the training set and select the most $K$ relevant examples $\mathcal{D}^e_i$:
\begin{equation}
    \mathcal{D}^e_i = \{(\boldsymbol{x}_j, \boldsymbol{y}_j) ~|~ \boldsymbol{x}_j \in \text{top} ~K (sim(\boldsymbol{x}_i, \boldsymbol{x}_j))\}
\end{equation}
Inspired by compression techniques, we propose instance-level examples pruning to leverage the performance gains while mitigating the generation of substantial additional overhead. We gradually reduce the number of examples from $K$ to $0$ over a total of $T$ schedule steps, to achieve complete example internalization. The number of examples $K^t$ at $t$-th step can be expressed as:
\begin{equation}
    K^t = \lfloor K \times \mathcal{S}_t \rfloor
\end{equation}
Finally, we randomly select $K^t$ examples from the set $\mathcal{D}^e_i$ as examples $\boldsymbol{e}_i^t$ for $t$-th step fine-tuning.


\subsection{SKIntern Pipeline}
\label{pipeline}

\noindent \textbf{Fine-tuning SLMs with Rationales.} For each specific schedule step $\mathcal{S}_t$, we utilize the compressed symbolic knowledge $\boldsymbol{k}_i^t$ and pruned examples $\boldsymbol{e}_i^t$ for fine-tuning the SLM $p_{\theta}$ with trainable parameters $\theta$ to generate the rationale $\boldsymbol{r}_{ij}$ and answer $\boldsymbol{y}_i$ for the question $\boldsymbol{x}_i$ as follows:
\begin{equation}
    \mathcal{L}_{t}(\theta) = - \frac{1}{n \cdot l} \sum_{i=1}^{n} \sum_{j=1}^{l} \log p_{\theta}(\boldsymbol{r}_{ij}, \boldsymbol{y}_i \mid \boldsymbol{k}_i^t, \boldsymbol{e}_i^t, \boldsymbol{x}_i)
\end{equation}
We aim to minimize the negative log-likelihood of the sequence comprising the rationale $\boldsymbol{r}_{ij}$ and answer $\boldsymbol{y}_i$, ensuring rationale precedes the answer.

\noindent \textbf{Progressive Fine-tuning.} For a total of $T$ schedule steps, we fine-tune the SLM parameters with the learning rate $\eta$ for internalizing as follows:
\begin{equation}
    \theta_{t + 1} = \theta_t - \eta \nabla_{\theta}\mathcal{L}_{t}(\theta)
\end{equation}

\noindent \textbf{Inference.} After progressive fine-tuning, we utilize the updated model parameters, denoted as $\theta_T$, to conduct inferences without the need for additional knowledge or examples. Consequently, we can simply handle the question and complete efficient and effective inference.

\section{Experiment}

In this section, we conduct extensive experiments and comprehensive analysis to evaluate the effectiveness of \textit{SKIntern} on both in-domain (ID) and out-of-domain (OOD) datasets.

\begin{table*}[t]
  \centering
  \resizebox{\linewidth}{!}{
      \begin{tabular}{lccccccccc}
        \toprule
         \multirow{2}{*}{\textbf{Methods}} & \multicolumn{2}{c}{\textbf{In-Domain}}& \multicolumn{5}{c}{\textbf{Out-Of-Domain}} & \multirow{2}{*}{\textbf{Avg}} & \textbf{Rel.}\\
        \cmidrule(r){2-3}\cmidrule(r){4-8}
         & \textbf{BBH-test} & \textbf{GSM8K} & \textbf{BB-sub} & \textbf{AGIEval} & \textbf{GSM8K-PLUS} & \textbf{ARC-E}  & \textbf{ARC-C} & & \textbf{FLOPs}  \\
        \midrule
        \rowcolor{gray!20} \multicolumn{10}{l}{\textit{\# Closed-source model and Open-source models (Zero-shot-CoT)}}\\
        GPT-3.5-turbo (\textit{Teacher}) & 43.2 & 72.6 & 44.0 & 50.5 & 55.9 & 91.8 & 84.1 & 63.2 & - \\
        LLaMA-3-70B-Instruct & 62.6 & 89.2 & 51.0 & 66.3 & 72.9 & 97.6 & 93.2 & 76.1 & -\\
        \midrule
        \rowcolor{gray!20} \multicolumn{10}{l}{\textit{\# TinyLLaMA-1.1B based}}\\
        Zero-shot \cite{zeroshot}        & 14.0 & 2.0 & 17.7 & 17.8 & 1.5 & 19.4 & 15.0 & 12.5 & $\times$1.0\\
        Zero-shot-CoT \cite{zeroshotcot} & 13.5 & 1.4 & 17.7 & 10.4 & 1.3 & 16.0 & 13.4 & 10.5 & $\times$1.0\\
        \hdashline
        Fine-tuning                     & 48.8 & 3.5 & 26.0 & 21.2 & 3.7 & 28.0 & 24.6 & 22.3 & $\times$\textbf{0.9}\\
        Knowledge-Augmented Fine-tuning & 49.3 & 3.7 & 27.4 & 21.9 & 3.3 & 29.4 & 25.3 & 22.9 & $\times$3.7\\
        \hdashline
        Std-CoT \cite{stdcot} & 47.8$_{\pm .43}$ & 7.9$_{\pm .27}$ & 27.6$_{\pm .31}$ & 21.5$_{\pm .56}$ & 4.3$_{\pm .62}$ & 28.2$_{\pm .69}$ & 25.0$_{\pm .48}$ & 23.2 & $\times$1.0\\
        MT-CoT \cite{mt-cot} & 44.1$_{\pm .78}$ & 4.1$_{\pm .35}$ & 25.0$_{\pm .45}$ & 21.4$_{\pm .64}$ & 2.8$_{\pm .83}$ & 33.5$_{\pm .52}$ & 25.1$_{\pm .59}$ & 22.3 & $\times$\textbf{0.9} \\
        Step-by-step \cite{distilling} & 42.4$_{\pm .56}$ & 4.3$_{\pm .47}$ & 26.2$_{\pm .38}$ & 21.1$_{\pm .72}$ & 3.1$_{\pm .54}$ & 29.6$_{\pm .61}$ & 25.9$_{\pm .66}$ & 21.8 & $\times$\textbf{0.9} \\
        KARD (BM25) \cite{kard} & 49.5$_{\pm .61}$ & 7.6$_{\pm .40}$ & 26.9$_{\pm .43}$ & 20.2$_{\pm .48}$ & 4.0$_{\pm .77}$ & 28.2$_{\pm .85}$ & 26.5$_{\pm .91}$ & 23.3 & $\times$3.9\\
        CasCoD \cite{improve} & 48.1$_{\pm .49}$ & 6.8$_{\pm .39}$ & 23.1$_{\pm .64}$ & 19.4$_{\pm .73}$ & 4.8$_{\pm .48}$ & 29.0$_{\pm .63}$ & 27.1$_{\pm .42}$ & 22.6 & $\times$3.0\\
        \textbf{SKIntern} (\textit{ours}) & \textbf{55.5$_{\pm .71}$} & \textbf{8.1$_{\pm .65}$} & \textbf{31.4$_{\pm .44}$} & \textbf{24.4$_{\pm .90}$} & \textbf{5.3$_{\pm .68}$} & \textbf{36.8$_{\pm .89}$} & \textbf{31.2$_{\pm .32}$} & \textbf{27.5} & $\times$1.0\\
        \midrule
        \rowcolor{gray!20} \multicolumn{10}{l}{\textit{\# LLaMA2-7B based}}\\
        Zero-shot \cite{zeroshot}        & 17.3 & 2.7 & 18.6 & 19.2 & 2.4 & 25.2 & 20.6 & 17.0 & $\times$6.4 \\
        Zero-shot-CoT \cite{zeroshotcot} & 13.5 & 3.1 & 12.2 & 10.3 & 2.1 & 29.1 & 20.2 & 12.9 & $\times$6.4 \\
        \hdashline
        Fine-tuning                     & 57.8 & 5.8 & 33.3 & 31.0 & 5.8 & 73.3 & 56.3 & 37.6 & $\times$\textbf{5.6}\\
        Knowledge-Augmented Fine-tuning & 58.7 & 6.3 & 34.2 & 31.8 & 6.1 & 75.1 & 57.0 & 38.5 & $\times$23.7\\
        \hdashline
        Std-CoT \cite{stdcot}  & 58.1$_{\pm.74}$ & 20.5$_{\pm.71}$  & 30.7$_{\pm.48}$ & 23.6$_{\pm.65}$    & 12.0$_{\pm.26}$   & 73.4$_{\pm.81}$  & 55.9$_{\pm.78}$  & 39.2 & $\times$6.4  \\
        MT-CoT \cite{mt-cot} & 45.6$_{\pm .43}$ & ~~6.8$_{\pm .59}$ & 27.8$_{\pm .75}$ & 31.7$_{\pm .89}$ & ~~6.0$_{\pm .72}$ & 74.2$_{\pm .46}$ & 57.6$_{\pm .38}$ & 35.7 & $\times$5.7\\
        Step-by-step \cite{distilling} & 54.3$_{\pm .37}$ & ~~8.4$_{\pm .93}$ & 32.9$_{\pm .55}$ & 32.4$_{\pm .64}$ & ~~5.9$_{\pm .57}$ & 77.7$_{\pm .35}$ & 61.8$_{\pm .87}$ & 39.1 & $\times$\textbf{5.6} \\
        KARD (BM25) \cite{kard} & 58.9$_{\pm .53}$ & 27.5$_{\pm .71}$ & 30.3$_{\pm .45}$ & 18.9$_{\pm .38}$ & 19.1$_{\pm .73}$ & 73.7$_{\pm .41}$ & 57.0$_{\pm .82}$ & 40.8 & $\times$24.5 \\
        CasCoD \cite{improve} & 58.9$_{\pm .59}$ & 29.2$_{\pm .75}$ & 32.2$_{\pm .36}$ & 28.8$_{\pm .29}$ & \textbf{21.4}$_{\pm .79}$ & 74.7$_{\pm .91}$ & 57.3$_{\pm .62}$ & 43.2 & $\times$19.0 \\
        \textbf{SKIntern} (\textit{ours}) & \textbf{69.3}$_{\pm.58}$ & \textbf{33.9}$_{\pm.71}$ & \textbf{37.2}$_{\pm.51}$ & \textbf{31.3}$_{\pm.49}$ & 21.2$_{\pm.83}$ & \textbf{78.1}$_{\pm.24}$ & \textbf{62.1}$_{\pm.67}$ & \textbf{47.6} & $\times$6.4 \\
        \bottomrule
      \end{tabular}
      }
    \caption{Performance (\%) of LLaMA2-7B \cite{llama} and TinyLLaMA-1.1B \cite{TinyLlamaAO} with different methods across seven selected datasets. \textbf{Bold} indicates the best in each setting. We report the mean and standard deviation of accuracy with 3 different runs for CoT distillation methods. Relative FLOPs cost is calculated relative to the TinyLLaMA with Zero-shot. We calculate the FLOPs required on BBH-test for each method.
    }
    \label{main}
\end{table*}

\subsection{Datasets}
\label{sec:dataset}

Following \citet{llms}, we focus on three practical abilities: factual, mathematical, and general reasoning. For each ability, we select a relevant public dataset as the ID dataset, integrate its training data into the target dataset \(\mathcal{D}_{\text{train}}\) for mixed training, and combine its test data into the evaluation dataset \(\mathcal{D}_{\text{eval}}\). Additionally, each ability includes OOD datasets in \(\mathcal{D}_{\text{eval}}\), allowing us to evaluate the model's ability to generalize and enhance performance beyond the ID training environment.

\noindent \textbf{Factual Reasoning}: We select the Multitask Language Understanding (MMLU) \cite{mmlu} as the ID dataset, which includes multiple-choice questions across 57 subjects. For OOD evaluation, we use the ARC \cite{arc}, comprising both Easy and Challenge segments.

\noindent \textbf{Mathematical Reasoning}: We select MetaMathQA \cite{metamath} as the ID dataset, which only has a training set that includes a high-quality collection of mathematical reasoning question-answer pairs, derived from GSM8K \cite{gsm8k} and MATH \cite{math}. We use GSM8K as the ID evaluation and GSM8K+ \cite{gsmplus} for OOD evaluation.

\noindent \textbf{General Complex Reasoning}: We chose BIG-Bench Hard (BBH) \cite{bbh} as the ID dataset, which includes 27 challenging tasks spanning arithmetic, symbolic reasoning, and more, derived from BIG-Bench (BB) \cite{bb}. Most of the data consists of multiple-choice questions. For OOD evaluation, we use BB-Sub filtered by CasCoD, and AGIEval \cite{agieval} subtasks about English multiple-choice questions.

\subsection{Baselines}

We compare our method with the following baselines: \textit{1)} \textbf{Teacher \& Vanilla Student} in Zero-shot \cite{zeroshot} and Zero-shot-CoT \cite{zeroshotcot}. \textit{2)} \textbf{Fine-tuning} involves fine-tuning a model to generate answers given only questions. The performance of the baselines above illustrates the capability of SLMs to solve tasks using only training data, without external guidance or additional knowledge. \textit{3)} \textbf{CoT distillation} includes \textbf{Std-CoT} \cite{stdcot} which is the standard CoT distillation method, enabling direct fine-tuning of the student model with CoT data; \textbf{Step-by-step} \cite{distilling} is a multi-task method that extracts rationales and answers separately; \textbf{MT-CoT} \cite{mt-cot} is another multi-task method that optimizes both answer prediction and CoT generation simultaneously; \textbf{CasCoD} \cite{improve} decomposes the traditional single-step learning process into two cascaded learning steps. \textit{4)} \textbf{Knowledge-Augmentation} involves attaching retrieved passages to the question during both training and inference. This includes \textbf{Knowledge-Augmented Fine-tuning} focuses on generating answers only, and \textbf{KARD} \cite{kard} emphasizes learning the generation of rationales.

\begin{figure}[t]
\centerline{\includegraphics[width=0.5\textwidth]{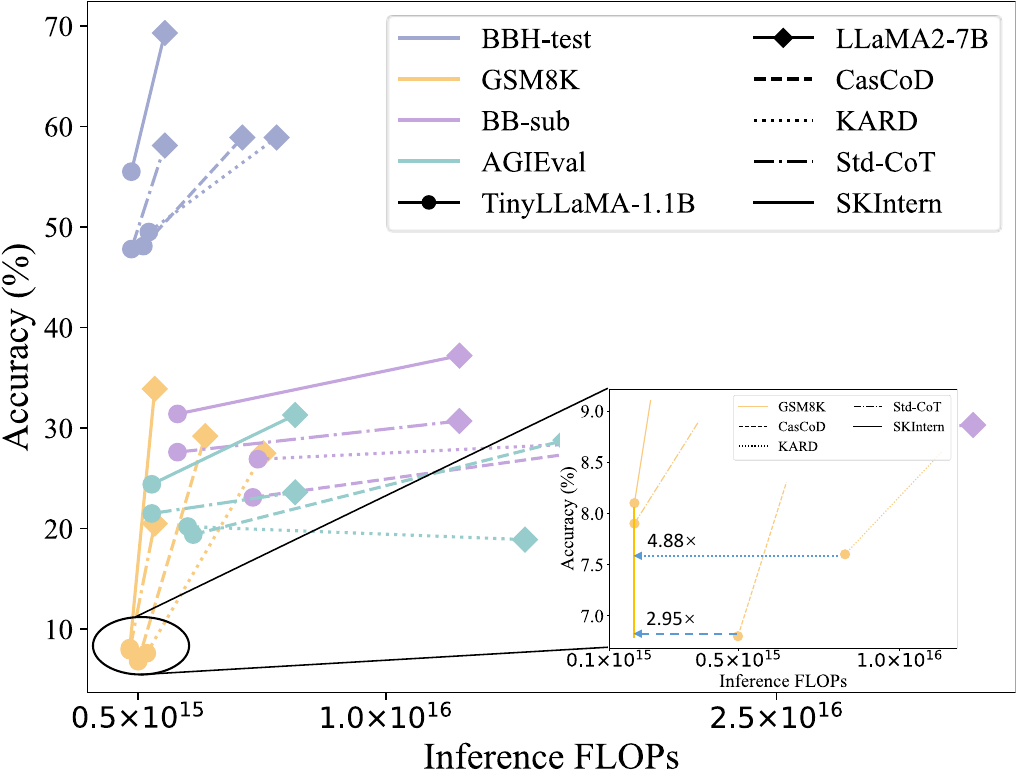}}
\caption{Accuracy (\%) against FLOPs for varying model sizes. FLOPs calculations are based on processing all examples from the same task during inference.
}
\label{flops}
\end{figure}

\begin{figure*}[t]
\centering
\includegraphics[width=1.8\columnwidth]{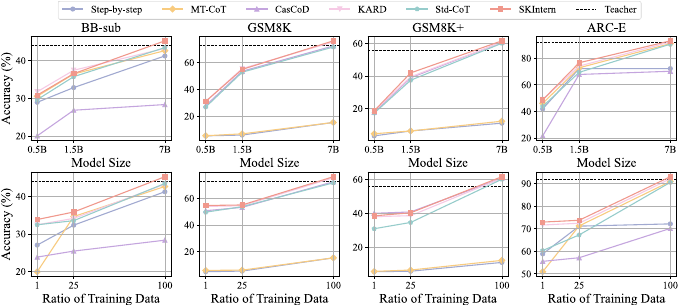} 
\caption{Efficiency on training data and model size. The backbone model for the data size variation is Qwen2-7B.}
\label{size}
\end{figure*}

\subsection{Implementations}

For all experiments, we use the LLaMA3-8B, LLaMA2-7B \cite{llama}, Qwen2 (0.5B, 1.5B, 7B) \cite{Qwen2TR} and TinyLLaMA-1.1B \cite{TinyLlamaAO} as the student SLM. We query the teacher model GPT-3.5-turbo to annotate the CoTs data with the manual prompt \cite{bbh}. Unless otherwise specified, $T$ is set to 4 (§\ref{sch}), and total epochs $E$ is set to 12.

We employ LoRA \cite{hu2022lora} for parameter-efficient fine-tuning of the student SLMs. All experiments are conducted on 2 A100 GPUs with 80GB. During the inference stage, we utilize vLLM \cite{vllm} to accelerate inference. Detailed information about training, inference and hyperparameters is provided in Appendix \ref{experiment}.

\subsection{Main Results}

We report the performance and inference costs of \textit{SKIntern} and baselines in
Table \ref{main} and Figure \ref{flops} (More results are shown in Appendix \ref{exre}) and find:


\textbf{\textit{SKIntern} outperform baselines with fewer FLOPs.} As shown in Figure \ref{flops}, when FLOPs-matched (in a vertical comparison), \textit{SKIntern} outperforms KARD which retrieves documents to augment reasoning, and CasCoD which enhances reasoning by cascaded decomposition. Specifically, from Table \ref{main}, it is evident that \textit{SKIntern} shows an average improvement of 8.4\% with LLaMA2-7B and 5.9\% with TinyLLaMA-1.1B, respectively. This highlights the utility of dynamic pruning and gradual internalization of symbolic knowledge.

\textbf{\textit{SKIntern} are up to $4\times$ more efficient than baselines.} Table \ref{main} demonstrates that \textit{SKIntern} uses 2-4$\times$ fewer FLOPs than state-of-the-art KARD and CasCoD. Although other CoT distillation methods can achieve similar computational savings, their performance is significantly worse than \textit{SKIntern} ($\geq$ 8\%). Specifically, their performance is 10\% lower on the mathematical reasoning dataset GSM8K and 15\% lower on the complex reasoning dataset BBH. Furthermore, \textit{SKIntern} achieves comparable performance with fewer FLOPs, as shown in Figure \ref{flops} (in a horizontal comparison).




\subsection{Efficiency on Dataset and Model Sizes}
\label{model_size}

To evaluate the efficiency of \textit{SKIntern} in terms of training data and model size, we measured test accuracy using Qwen2 \cite{Qwen2TR} models across various methods while varying the amount of training data and model size. As shown at the bottom of Figure \ref{size}, \textit{SKIntern} successfully transfers the reasoning ability of the teacher LLM into the parameters, even with minimal training data. As the amount of training data increases, \textit{SKIntern} consistently outperforms other baselines, with the improvement magnitude growing as well. This suggests that \textbf{\textit{SKIntern} performs optimally across different data volumes and achieves superior reasoning ability distillation}. Even with a limited dataset, \textit{SKIntern} outperforms other methods, demonstrating robustness and sample efficiency.

Regarding model size efficiency, as shown at the top of Figure \ref{size}, \textit{SKIntern} outperforms other baselines across various model scales. Notably, \textit{SKIntern} enables Qwen2-7B to surpass the teacher model, GPT-3.5 Turbo, in both ID and OOD tasks, despite having fewer parameters. \textit{SKIntern} offers substantial advantages for models of varying sizes, consistently outperforming other methods. These results underscore the practical benefits of \textit{SKIntern} in resource-limited environments, as it reduces the computational demands for SLMs while delivering performance on par with or surpassing larger models. \textbf{This further demonstrates that SLMs (0.5B) struggle to fully leverage CoT reasoning generated by LLMs, highlighting the need for our \textit{SKIntern} approach.}

\begin{table}[t]
\centering
\renewcommand\arraystretch{1.05}
\resizebox{\linewidth}{!}{
\begin{tabular}{lccccc}
\hline
\textbf{SKIntern} & \textbf{BBH} & \textbf{BB} & \textbf{AGIEval} & \textbf{GSM8K+} & \textbf{ARC-E} \\
\hline
\rowcolor{gray!20} \multicolumn{6}{l}{\textit{Pattern of Schedule $\mathcal{S}$}}\\
~~- exp & 64.8  & 36.2 & 30.0 & 16.3 & 76.0 \\
~~- exp$^{-1}$ & 59.5  & 31.2 & 28.8 & 15.4 & 73.9 \\
~~- linear & \textbf{69.3}  & \textbf{37.2} & \textbf{31.3} & \textbf{21.2} & \textbf{78.1} \\
\hline
\rowcolor{gray!20} \multicolumn{6}{l}{\textit{Step of Schedule $T$}}\\
~~- $T=3$ & 60.2  & 33.4 & 29.1 & 15.5 & 74.8 \\
~~- $T=4$ & \textbf{69.3}  & \textbf{37.2} & \textbf{31.3} & \textbf{21.2} & \textbf{78.1} \\
~~- $T=7$ & 65.7  & 35.0 & 30.0 & 20.9 & 76.6 \\
\hline
\end{tabular}
}
\caption{Comparison of schedule patterns and steps of \textit{SKIntern}. The backbone model is LLaMA2-7B.}
\label{table:pattern}
\end{table}


\subsection{Analysis on Schedule}
\label{sch}

\noindent \textbf{Schedule Pattern.} We examine the effectiveness of different schedule patterns during the progressive fine-tuning process, focusing on their impact on reasoning performance. The patterns tested include exponential, inverse exponential, and linear decay. As shown in Table \ref{table:pattern}, the linear decay consistently delivers the highest performance
, showcasing superior parsing efficiency and language understanding. In contrast, the inverse exponential schedule exhibits the lowest effectiveness, while the exponential decay offers moderate performance but remains inferior to the linear schedule. These findings indicate that \textbf{a gradual, steady reduction is more advantageous than a more aggressive approach}. Progressive fine-tuning with a linear decay schedule appears to yield optimal performance compared to other patterns.

\noindent \textbf{Schedule Setup.} We explore the optimal schedule step $T$ for linear decay during progressive fine-tuning. With the total number of epochs set to 12, we chose the common divisors of 12 for linear decay, where $T$ corresponds to the decay step plus 1. As seen in Table \ref{table:pattern}, $T=4$ offers the best performance, while $T=7$ shows slightly lower results, and $T=3$ yields the poorest performance. This suggests that overly frequent schedule changes hinder sufficient learning in the initial stages, whereas sparse schedules cause large, disruptive jumps, complicating smooth progression and increasing learning difficulty. Therefore, \textbf{selecting an appropriate schedule step is crucial for effectively internalizing knowledge and enhancing reasoning abilities in SLMs}.

\subsection{Ablation Studies}
\label{ab}

To demonstrate the effectiveness of \textit{SKIntern}, we conducted ablation studies using LLaMA2-7B by creating three variants: (1) w/o \(\boldsymbol{k}^m\), which removes the learning summary during fine-tuning; (2) w/o \(\boldsymbol{k}^p\), where supplemental knowledge is excluded; and (3) w/o \(\boldsymbol{e}\), where example pruning is omitted. As shown in Table \ref{table:ab}, the removal of any of these components results in reduced performance, highlighting the critical role of internalizing both knowledge and examples in enhancing SLMs' complex reasoning abilities during progressive fine-tuning.


\begin{table}[t]
\centering
\renewcommand\arraystretch{1.05}
\resizebox{\linewidth}{!}{
\begin{tabular}{lccccc}
\hline
\textbf{Methods} & \textbf{BBH} & \textbf{BB} & \textbf{AGIEval} & \textbf{GSM8K+} & \textbf{ARC-E} \\
\hline
\textbf{SKIntern} & \textbf{69.3}  & \textbf{37.2} & \textbf{31.3} & \textbf{21.2} & \textbf{78.1} \\
w/o $\boldsymbol{k}^m$ &  59.8 & 30.8 & 28.7 & 15.3 & 74.1 \\
w/o $\boldsymbol{k}^p$ & 62.3 & 32.1 & 29.5 & 16.2 & 75.7 \\
w/o $\boldsymbol{e}$ & 61.9 & 34.1 & 29.4 & 18.1 & 74.6\\
\hline
\end{tabular}
}
\caption{Ablation studies on different components.}
\label{table:ab}
\end{table}

\begin{figure}[t]
\centerline{\includegraphics[width=0.42\textwidth]{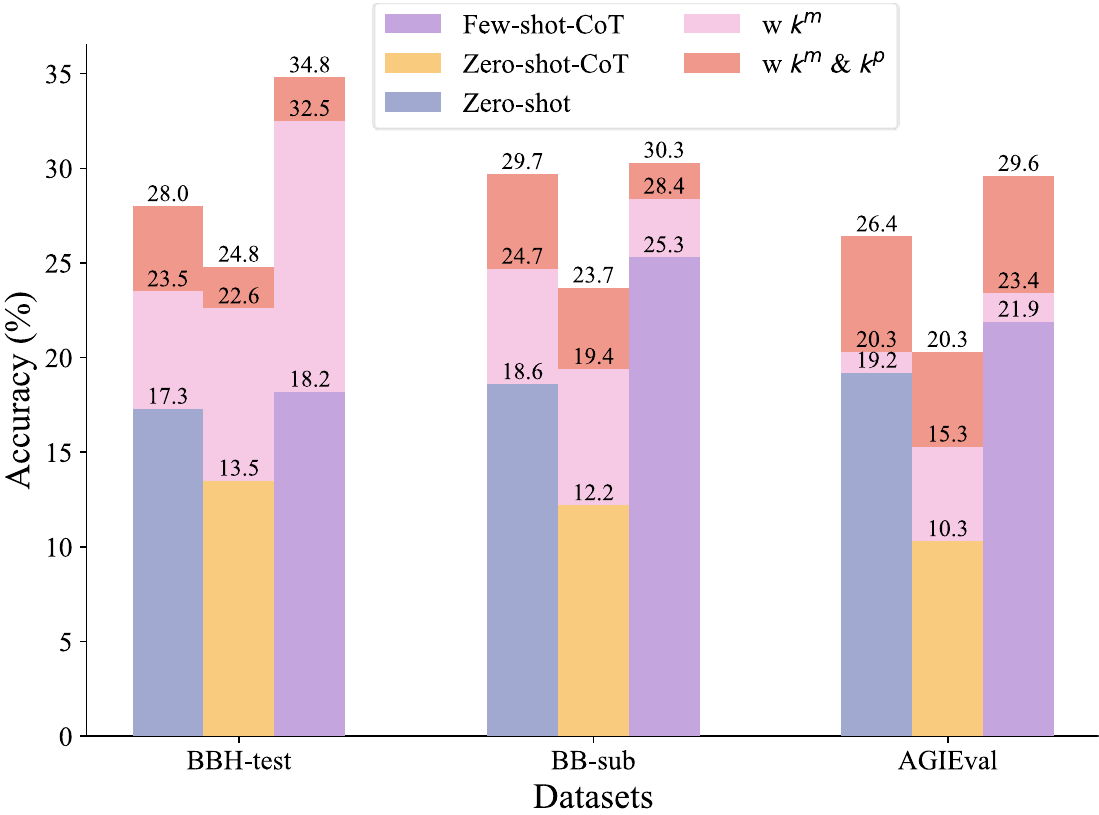}}
\caption{Ablation studies of $\boldsymbol{k}$ on vanilla methods.}
\label{fig4}
\end{figure}

Additionally, we investigate the effectiveness of the generated symbolic knowledge (see Figure \ref{fig4}). Incorporating learning summaries $\boldsymbol{k}^m$ and supplementary knowledge $\boldsymbol{k}^p$ into the original zero-shot, zero-shot-cot, and few-shot-cot significantly enhances performance. Remarkably, this improvement occurs without fine-tuning, demonstrating the utility and generalization of symbolic knowledge in augmenting the model's inference capabilities.

\section{Conclusion}

In this paper, we introduce \textit{SKIntern}, a novel CoT distillation method designed to internalize symbolic knowledge and rich examples into model parameters, thereby enhancing the ability of SLMs to tackle complex reasoning tasks. Through a systematic schedule, symbolic knowledge generated by the LLM including learning summaries and supplementary knowledge is compressed and selected examples are refined. These elements are then used to fine-tune the SLM, enabling it to produce coherent rationales and accurate answers. We implement a customized progressive fine-tuning pipeline to accommodate various schedule steps and training epochs. Extensive experiments demonstrate that our method not only improves reasoning performance on both in-domain (ID) and out-of-domain (OOD) tasks but also significantly accelerates inference and reduces computational resource usage.

\section*{Limitations}

\noindent \textbf{Method} We have demonstrated through \textit{SKIntern} that the performance of SLM on complex inference tasks can be significantly improved while greatly reducing computational overhead. However, it is important to acknowledge the limitations of our research. The effectiveness of our knowledge enhancement largely depends on the incremental fine-tuning required to internalize the original symbolic knowledge and examples, which increases the complexity and cost of training. Additionally, using LLM to generate supplementary symbolic knowledge necessitates further monetary expenditure due to API calls.

\noindent \textbf{Task} While our current tests encompass factual knowledge, mathematics, and complex reasoning, the method's efficacy for different tasks, such as various coding exercises and extended text tasks, requires further analysis and experimentation. Additionally, further investigation is needed to determine which types of symbolic knowledge and task examples are more easily learned and internalized.

\noindent \textbf{Large Language Models} Regarding the experiments, given our limited computing and financial budgets, we chose GPT-3.5-Turbo as the teacher. Using GPT-4 would likely better verify the effectiveness of our method, \textit{SKIntern}. Additionally, our aim to enhance the complex reasoning ability of SLMs restricted our choice to mainstream models, such as Llama2, Llama3, and Qwen2, thereby excluding other excellent models like Phi3 and DeepSeek. However, exploring larger LMs such as 13B and 72B with \textit{SKIntern} could be of great interest, presenting a promising direction for future research. Experimental results indicate that enhancing powerful models like Llama3-8B and Qwen2-7B surpasses GPT-3.5-Turbo and matches Llama3-70B.

\section*{Ethical Considerations}
In this paper, we proposed a novel knowledge enhancement method aimed at leveraging the knowledge of LLMs. However, LLMs may generate inappropriate or discriminatory knowledge. Our approach does not introduce ethical concerns. The datasets we used are public, and there are no privacy issues.
\section*{Acknowledgements}

This work was supported by the National Key R\&D Program of China (No. 2022ZD0160503) and the National Natural Science Foundation of China (No.62376270, No.62276264).
\bibliography{custom}

\appendix

\section{Experimantal Settings}
\label{experiment}

\subsection{Datasets}
\label{app_data}

For each ability, we select a relevant public dataset, integrate its training data into the target dataset \(\mathcal{D}_{\text{train}}\) for mixed training, and combine its test data into the evaluation dataset \(\mathcal{D}_{\text{eval}}\). Additionally, each ability includes an OOD dataset in \(\mathcal{D}_{\text{eval}}\). This setup allows us to evaluate the model's ability to generalize and enhance performance beyond the ID training environment.

Table \ref{stat_data} shows the statistics details of the selected datasets.

\begin{table*}[htbp]
  \centering
  \renewcommand\arraystretch{1.05}
      \begin{tabular}{llccc}
        \toprule
        \textbf{Abilities} & \textbf{Task} & \textbf{\# Train} & \textbf{\# Train (Filtered)} & \textbf{\# Test} \\
        \midrule
        \multirow{3}{*}{Factuality} & ID: MMLU & Dev + Val: 1,815 & 1,555 & - \\
                                    & OOD: ARC-C & -  & - & 1,172 \\
                                    & OOD: ARC-E & - & - & 2,376 \\
                                    \hline
        \multirow{3}{*}{Math} & ID: MetaMathQA & 395,000  & 3,500 & - \\
                                    & OOD: GSM8K & -  & - & 1,319 \\
                                    & OOD: GSM8K-PLUS & - & - & 1,400 \\
                                    \hline
        \multirow{3}{*}{Reasoning} & ID: BBH & 6,511  & 3,805 & 1,304 \\
                                    & OOD: BB-sub & -  & - & 5,384 \\
                                    & OOD: AGIEval & - & - & 2,546 \\
                                    \hline
        \textbf{All} & \textbf{Sum} & - & 8,860 & 15,501\\
        \bottomrule
      \end{tabular}
    \caption{Statistical details of the selected datasets. Since MMLU lacks official training data, we combined the development and validation datasets to form a training set. To maintain sample balance, we matched the size of MetaMathQA to that of BBH. We obtained balanced samples from two dataset augmentation modes, MATH\_Aug and GSM\_Aug, resulting in a total of 3,500 samples.}
    \label{stat_data}
\end{table*}

For MMLU \cite{mmlu}, we adhere to previous prompt styles \cite{bbh}, instructing the teacher model (e.g., GPT-3.5-Turbo) to generate answers and Chains of Thought (CoT). By excluding samples with incorrect answers, we ultimately obtained a total of 1,556 samples. For MetaMathQA \cite{metamath}, we acquired 3,500 samples through random sampling. For BB \cite{bb}, we followed the CasCoD \cite{improve} methodology by filtering the original dataset for tasks containing the keyword "multiple choice" and randomly extracting up to 100 examples for each task. Note that tasks in BBH do not involve BB-sub. 

During the evaluation stage, we use Exact Match \cite{squad} as the evaluation metric.

The answer generation between the involved models is conducted in a zero-shot setting, with all models set to a temperature of 0.8 and a maximum token length of 1024. The prompt can be found in the Appendix \ref{prompt_cot}.

\begin{table*}[htbp]

  \centering
  \renewcommand\arraystretch{1.05}
   \resizebox{\linewidth}{!}{
      \begin{tabular}{lcccccc}
        \toprule
        \textbf{Hyperparameter}  & \textbf{TinyLLaMA-1.1B} &  \textbf{LLaMA2-7B} & \textbf{LLaMA3-8B}& \textbf{Qwen2-0.5B}& \textbf{Qwen2-1.5B}& \textbf{Qwen2-7B}\\
        \midrule
        Max Input Len & 2048 & 4096 & 4096 & 4096 & 4096 & 4096 \\
        Max Output Len & 128 & 128 & 128 & 128 & 128 & 128  \\
        Optimizer & AdamW & AdamW  & AdamW & AdamW & AdamW & AdamW \\
        Learning Rate & 2e-4 & 1e-4  & 5e-5 & 2e-4 & 1e-4 & 1e-4 \\
        precision & fp16 & fp16 & fp16 & fp16 & fp16 & fp16 \\
        \# Training epochs & 12 & 12 & 12  & 12 & 12 & 12 \\
        \# Warmup Steps & \multicolumn{6}{c}{10\% of total training steps}  \\
        Batch Size & 32 & 16 & 8 & 32 & 16 & 8  \\
        Gradient Accumulation & 1 & 2 & 4 & 1 & 2 & 4 \\
        rank of LoRA & 32 & 32 & 32 & 32 & 32 &32 \\
        \bottomrule
      \end{tabular}
      }
        \caption{Training hyperparameters.}
          \label{hyperparam}
\end{table*}

\begin{table}[htbp]

  \centering
  \renewcommand\arraystretch{1.05}
   \resizebox{\linewidth}{!}{
      \begin{tabular}{lccc}
        \toprule
        \multirow{2}{*}{\textbf{Hyperparameter}} & \multirow{2}{*}{\textbf{Student}} & \multicolumn{2}{c}{\textbf{Teacher}}\\
        \cmidrule(r){3-4}
        & & Rationale & Reasoning\\
        \midrule
        do\_sample & False & True & False\\
        temperature & 0.6 & 0.8 & 0.6 \\
        top-p & 0.95 & 1.0 & 0.95 \\
        top-k & 50 & 50 & 50\\
        max\_new\_tokens & 1024 & 2048 & 1024 \\
        \# return sequences & 1 & 2 & 1 \\
        \bottomrule
      \end{tabular}
      }
        \caption{Generation configs of students and teachers.}
          \label{gen}
\end{table}

\subsection{Hyperparameter}
\label{hyperp}
The complete set of stable hyperparameters used for both baseline models and the proposed \textit{SKIntern} training and inference runs can be found in Table \ref{hyperparam} and Table \ref{gen}, respectively.

In our research, we ensured consistent hyperparameter settings across all baselines, including the proposed \textit{SKIntern} method, to maintain the fairness of our comparative analysis. Detailed hyperparameters and their explanations are presented below. For \textit{SKIntern}, particularly in the fourth step, we reduced the enhanced distillation parameters to 3 epochs and fixed the batch size at 8, as the concatenation of specialized knowledge results in longer inputs. We maintained a consistent batch size across all baselines to eliminate any performance differences attributable to varying batch sizes, which depend on model size, with larger models use smaller batch sizes. The learning rate, a key parameter affecting model performance, was set to 5e-5, 1e-4, 2e-4, and 3e-4 in a series of experiments, revealing that larger models require smaller learning rates. Consequently, we adjusted the learning rate according to model size.

\subsection{Implementations}
\label{sec:imp}
Our implementations are based on huggingface transformers v4.42.1 \citep{transformers} using PyTorch v2.3.1 \citep{pytorch} and LlamaFactory \cite{llamafactory}.

For CasCoD \cite{improve}, we adhere to the optimal settings recommended by the authors, specifically setting $\alpha$ to 0.3. For KARD \cite{kard}, we employ the BM25 configuration \cite{bm25}, a sparse retrieval method based on word frequency, and retrieve three documents per question. Wikipedia serves as the external knowledge base for all datasets. For all retrievers used in \textit{SKIntern}, including BM25, Contriever \cite{contriever}, and DPR \cite{DensePR}, we utilize the Pyserini\footnote{https://github.com/castorini/pyserini} library, which offers a reproducible information retrieval framework.

\subsection{Symbolic Knowledge Collection}
\label{app_know}

For specialized knowledge collection, using 2-shot hand-written examples, the teacher model is configured with a temperature of 0.8 and a maximum length of 1024 tokens. It generates specialized knowledge corresponding to each incorrect example produced by the student SLMs. The prompt can be found in the Appendix \ref{prompt_know}.

\section{Extended Results}
\label{exre}
In Table \ref{ext}, we present the results of various models discussed in this paper, including LLaMA3-8B, QWen2-0.5B, 1.5B, and 7B, utilizing different baseline methods along with the outcomes of \textit{SKIntern}.

\begin{table*}[t]
  \centering
  \renewcommand\arraystretch{1.05}
  \resizebox{\linewidth}{!}{
      \begin{tabular}{lccccccccc}
        \toprule
         \multirow{2}{*}{\textbf{Methods}} & \multicolumn{2}{c}{\textbf{In-Domain}}& \multicolumn{5}{c}{\textbf{Out-Of-Domain}} & \multirow{2}{*}{\textbf{Avg}} & \textbf{Rel.}\\
        \cmidrule(r){2-3}\cmidrule(r){4-8}
         & \textbf{BBH-test} & \textbf{GSM8K} & \textbf{BB-sub} & \textbf{AGIEval} & \textbf{GSM8K-PLUS} & \textbf{ARC-E}  & \textbf{ARC-C} & & \textbf{FLOPs} \\
        \midrule
        \multicolumn{10}{l}{\textit{\# Closed-source model and Open-source models (Zero-shot-CoT)}}\\
        GPT-3.5-turbo (\textit{Teacher}) & 43.2 & 72.6 & 44.0 & 50.5 & 55.9 & 91.8 & 84.1 & 63.2 & - \\
        LLaMA-3-70B-Instruct & 62.6 & 89.2 & 51.0 & 66.3 & 72.9 & 97.6 & 93.2 & 76.1 & -\\
        \midrule 
         \rowcolor{gray!20} \multicolumn{10}{l}{\textit{\# LLaMA-3-8B based}}\\
        Zero-shot \cite{zeroshot}   & 18.2      & 2.8  & 27.4  & 29.7   & 2.2   & 50.8  & 50.0 &  25.9 & $\times$6.2 \\
        Zero-shot-CoT \cite{zeroshotcot} & 26.5 & 6.6  & 23.5 & 32.2    & 3.7   & 68.1  & 55.5  & 30.9 & $\times$6.2\\
        \hdashline 
        Fine-tuning & 43.7 & 11.7  & 29.1 & 35.3    & 9.4   & 75.2  & 65.2 & 38.5 & $\times$\textbf{5.4} \\
        Knowledge-Augmented Fine-tuning & 30.4 & 9.9  & 14.4 & 13.0    & 8.5   & 40.8  & 33.9  & 21.6 & $\times$23.3 \\
        \hdashline
        Std-CoT \cite{stdcot}  & 79.4 & 61.6  & 40.5 & 41.3    & 45.6   & 83.2  & 71.9  & 60.5 & $\times$6.2  \\
        MT-CoT \cite{mt-cot}   & 62.8 & 13.1  & 36.3 & \textbf{43.9}    & 11.4    & 83.6  & 72.3  & 46.3 & $\times$5.5  \\ 
        Step-by-step \cite{distilling}    & 64.0 & 11.5  & 38.8 & 43.7    & 9.0    & 84.3  & 74.6  & 46.6 & $\times$\textbf{5.4}  \\
        KARD (BM25) \cite{kard}     & \textbf{81.4} & \textbf{64.3}  & 43.1 & 43.4    & \textbf{48.6}   & 85.6  & \textbf{76.1}  & 63.2  & $\times$24.2 \\ 
        CasCoD \cite{improve}  & 32.1    & 59.1     & 18.1    & 23.6    &46.1   & 34.6     & 27.7  & 34.5 & $\times$17.7 \\ 
        \textbf{SKIntern} (\textit{ours}) & 80.8 & 62.5  & 42.8 & 43.6    & 48.1   & \textbf{89.9}  & 75.9  & \textbf{63.4} & $\times$6.2  \\
        \midrule
        \rowcolor{gray!20}\multicolumn{10}{l}{\textit{\# Qwen2-0.5B based}}\\
        Std-CoT \cite{stdcot}  & 65.8 & 26.7 & 29.6 & 25.6 & 17.1 & 43.6 & 32.0  & 34.3  & $\times$\textbf{0.4} \\
        MT-CoT \cite{mt-cot}   & 47.2 & 5.3 & 30.5 & \textbf{27.7} & 4.4 & 46.0 & 35.1  & 28.0  & $\times$\textbf{0.4} \\ 
        Step-by-step \cite{distilling}    & 44.2 & 5.2 & 28.9 & 26.2 & 3.1 & 41.8 & 36.2  & 26.5 & $\times$\textbf{0.4}  \\
        KARD (BM25) \cite{kard}     & \textbf{66.3} & \textbf{30.9} & 31.7 & 23.9 & 18.2 & \textbf{48.9} & \textbf{37.2}  & \textbf{36.7} & $\times$1.7  \\ 
        CasCoD \cite{improve}  & 37.6 & 27.7 & 20.0 & 15.6 & 17.6 & 21.5 & 14.8  & 22.1  & $\times$1.2 \\ 
        \textbf{SKIntern} (\textit{ours}) & 65.9 & \textbf{30.9} & \textbf{30.8} & 27.0 & \textbf{18.5} & 48.5 & 35.6  & \textbf{36.7} & $\times$\textbf{0.4}\\
        \midrule
        \rowcolor{gray!20}\multicolumn{10}{l}{\textit{\# Qwen2-1.5B based}}\\
        Std-CoT \cite{stdcot}  & 68.2 & 52.7 & 35.7 & 34.0 & 37.3 & 69.3 & 56.4  & 50.5  & $\times$1.3 \\
        MT-CoT \cite{mt-cot}   & 58.0 & 6.7 & 36.4 & 34.2 & 6.1 & 72.7 & 57.5  & 38.8  & $\times$\textbf{1.1} \\ 
        Step-by-step \cite{distilling}    & 48.4 & 5.8 & 32.8 & 34.4 & 6.1 & 72.1 & 57.6  & 36.7 & $\times$\textbf{1.1}  \\
        KARD (BM25) \cite{kard}     & \textbf{72.2} & \textbf{55.4} & \textbf{37.4} & 31.2 & 39.4 & 74.0 & 62.2 & 53.1  & $\times$5.2\\ 
        CasCoD \cite{improve}  & 31.7 & 53.4 & 25.4 & 24.7 & 38.8 & 57.1 & 47.8  & 39.8 & $\times$3.8  \\ 
        \textbf{SKIntern} (\textit{ours}) & 70.1 & 54.8 & 36.5 & \textbf{36.3} & \textbf{41.8} & \textbf{76.5} & \textbf{62.7}  & \textbf{54.1} & $\times$1.3 \\
        \midrule
        \rowcolor{gray!20}\multicolumn{10}{l}{\textit{\# Qwen2-7B based}}\\
        Std-CoT \cite{stdcot}  & \textbf{80.7} & 71.5 & 43.4 & \textbf{49.9} & 60.0 & 90.5 & 80.3 & 68.0 & $\times$6.0  \\
        MT-CoT \cite{mt-cot}   & 70.0 & 15.2 & 42.6 & 49.4 & 12.1 & 90.9 & 80.2  & 51.5  & $\times$5.3 \\ 
        Step-by-step \cite{distilling}    &  68.8 & 15.2 & 41.2 & 49.1 & 10.9 & 72.1 & 71.8  & 47.0 & $\times$\textbf{5.2}  \\
        KARD (BM25) \cite{kard}    & 80.2 & 75.3 & 43.2 & 49.6 & 60.6 & 92.1 & \textbf{83.5} & 69.2  & $\times$23.6   \\ 
        CasCoD \cite{improve}  & 35.7 & 72.3 & 23.8 & 37.4 & 60.6 & 70.1 & 63.1  & 51.9 & $\times$17.4 \\ 
        \textbf{SKIntern} (\textit{ours}) & 78.8 & \textbf{76.1} & \textbf{45.2} & 49.4 & \textbf{61.5} & \textbf{93.0} & \textbf{83.5}  & \textbf{69.6} & $\times$6.0 \\
        \bottomrule
      \end{tabular}
      }
    \caption{Performance (\%) of LLaMA3-8B \cite{llama} and Qwen2-0.5B, 1.5B and 7B \cite{Qwen2TR} with different methods across seven selected datasets. \textbf{Bold} indicates the best in each setting. Relative FLOPs cost is calculated relative to the TinyLLaMA with Zero-shot. We calculate the FLOPs required on BBH-test for each method.}
    \label{ext}
\end{table*}

\section{Case Study}
\label{case}

We present two cases from Tables \ref{tab:gsm} and \ref{tab:bbh} to compare the Chains of Thought (CoTs) generated by \textit{SKIntern}, the teacher large language model (LLM), and the standard CoTs distillation method (Std-CoT). We use \textcolor{green}{\ding{51}} and \textcolor{red}{\ding{55}} to indicate the correctness of the CoT. 

Table \ref{tab:gsm} shows that the Std-CoT's response is confused and fails to comprehend the question accurately. Although it has a rough idea, its rationale is entirely incorrect as it struggles to emulate the rationale of the teacher LLM. 

Table \ref{tab:bbh} presents the symbolic knowledge generated by the LLM for a training example in BBH, encompassing learning summaries and supplementary information. This symbolic knowledge offers detailed logical reasoning and positional insights, which assist the LLM in understanding and solving these problems.

\begin{table*}[htbp]
\centering
\renewcommand\arraystretch{1.15}
\resizebox{\linewidth}{!}{
\begin{tabular}{lp{8cm}p{8cm}}
\hline
\multicolumn{3}{l}{\makecell[l]{Question: Kylar went to the store to buy glasses for his new apartment. One glass costs \$5, but every second glass costs\\ only 60\% of the price. Kylar wants to buy 16 glasses. How much does he need to pay for them?}}
\\
\hline
 & \textbf{Symbolic Knowledge} & \textbf{Output} \\
\hline
\textbf{Teacher} & - & The discount price of one glass is 60/100 * 5 = \$<<60/100*5=3>>3. If every second glass is cheaper, that means Kylar is going to buy 16 / 2 = <<16/2=8>>8 cheaper glasses. So for the cheaper glasses, Kylar is going to pay 8 * 3 = \$<<8*3=24>>24. And for the regular-priced glasses, Kylar will pay 8 * 5 = \$<<8*5=40>>40. So in total Kylar needs to pay 24 + 40 = \$<<24+40=64>>64 for the glasses he wants to buy. Therefore, the answer is 64. ~~~ ~~~ ~~~\textcolor{green}{\ding{51}}\\
\hline
\textbf{Std-CoT} & - & 3.0 times a year means Pauline goes to the museum 3.0 * 5.0 = 15.0 times in 5.0 years. The cost of one visit is \$2.00, but in the last 5.0 years, it has increased by 1.5 times, so \textcolor{red}{the new cost is \$2.00 * 1.5 = \$3.00}. So, \textcolor{red}{Pauline spent 15.0 * 3.0 = \$45.00 on all visits to the museum in dollars}. Therefore, the answer is 45.  ~~~ ~~~ ~~~\textcolor{red}{\ding{55}}\\
\hline
\textbf{KARD} & 1. Conversely, deadweight loss can come from consumers if they buy a product even if it costs more than it benefits them. To describe this, if the same nail market had the government giving a \$0.03 subsidy to every nail produced, the subsidy would push the market price of each nail down to \$0.07(...) 2. Several glasses are utilized in transparent armor, such as normal plate glass (soda-lime-silica), borosilicate glasses, and fused silica. Plate glass has been the most common glass used due to its low cost(...) 3. To be eligible for the program, applicants must be in financial need, have had a recent eye exam, and have no other resources available to them to pay for glasses, including federal or state programs or assistance from local charitable organizations(...) & The cost of one glass is \$5. Kylar wants to buy 16 glasses, so he needs to \textcolor{red}{pay for 16 glasses + 1 glass = 17 glasses}. Since every second glass costs 60\% of the price, the cost of every second glass is 60\% of \$5 = \$3. So, the total cost of \textcolor{red}{the 17 glasses is \$5 x 17 + \$3 x (17/2) = \$85 + \$10}. Therefore, Kylar needs to pay \$85 + \$10 = \$95 for the glasses. Therefore, the answer is 95.  ~~~ ~~~ ~~~\textcolor{red}{\ding{55}}\\
\hline
\textbf{\textit{SKIntern}} & - & 16 glasses is an even number, so every second glass will cost 60\% of the price. Kylar wants to buy 16 glasses, and every second glass costs 60\% of the price, so he needs to buy 16/2 = 8 glasses at the full price of \$5/glass. He also needs to buy 8 glasses at the reduced price of 60\% of \$5/glass, which is 60/100 * \$5 = \$3/glass. So in total, Kylar needs to pay 8 * \$5 + 8 * \$3 = \$40 + \$24 = \$64 for the glasses. Therefore, the answer is 64.  ~~~ ~~~ ~~~\textcolor{green}{\ding{51}}\\
\hline
\end{tabular}}
\caption{A inference case from GSM8K.}
\label{tab:gsm}
\end{table*}

\begin{table*}[t]
\centering
\renewcommand\arraystretch{1.15}
\resizebox{\linewidth}{!}{
\begin{tabular}{p{16cm}}
\hline
\multicolumn{1}{l}{\makecell[l]{Question: If you follow these instructions, do you return to the starting point?\\ Take 7 steps. Take 9 steps. Turn around. Turn right.\\Options:\\- Yes\\- No}}
\\
\hline
\textbf{Symbolic Knowledge} \\
\hline
\textbf{Learning Summary}: When solving logical reasoning problems involving directions and movements, it's essential to follow the steps methodically and visualize the movements if necessary. 1. Question Comprehension: Begin by understanding the sequence of movements and turns. Identify the starting point and note the instructions given, such as steps taken and direction changes. 2. Thought Steps: - Start by taking note of the initial position and direction. - Follow the movement instructions step by step: - Take 7 steps in the initial direction. - Take 9 steps in the same direction. - Turn around (180-degree turn) which reverses your direction. - Turn right, which changes your direction perpendicularly. - After executing these steps, assess whether you return to the original position or direction. 3. Visualization: Drawing a simple diagram or using a grid can help track the positions and directions. This visualization helps verify whether the initial and final positions match. 4. Summative Experience: For similar questions, always track each movement and turn carefully. Be aware of the effects of each instruction, particularly turns, which change direction.\\ 
\textbf{Supplementary Knowledge}: 1. Understanding Directions: - Familiarize yourself with basic directions (e.g., north, south, east, west) and understand relative turns (left, right, and turn around). - A 180-degree turn changes direction to the opposite, while a 90-degree right or left turn changes the direction perpendicularly. 2. Visualization Techniques: - Use diagrams, sketches, or grids to map directions and movements to see the path clearly. - Visual aids can help prevent confusion, especially when multiple turns are involved. 3. Logical Sequencing: - Carefully follow each step in the sequence as instructed. Misinterpreting a step or turn can lead to incorrect conclusions. - Practice breaking down instructions into smaller parts to manage them more effectively. 4. Definitions: - Turn Around: A 180-degree turn where you face the opposite direction from where you started. - Right Turn: A 90-degree turn to the right, changing the direction perpendicular to the current path. By practicing these steps and understanding the underlying concepts, students can improve their ability to solve similar direction-based logical reasoning problems. \\
\hline
\end{tabular}}
\caption{A symbolic knowledge generation case from BBH-test.}
\label{tab:bbh}
\end{table*}

\section{Instruction Details}

\subsection{Prompt for Generating CoTs}
\label{prompt_cot}

We use the prompt template below to call the teacher model to generate the CoTs for the training datasets.

    ~ 
\begin{tcolorbox}
    [width=\textwidth, title = {Generate CoTs}]
    
    You are an expert assistant teacher. The following are tasks about \{Task\_Name\}. \{Task Description\}. Explain your reasoning first and your response should conclude with the format ``Therefore, the answer is".

    ~ 
    
    Question: \{QUESTION\}
    
    Answer: Let’s think step by step.
\end{tcolorbox}

\subsection{Prompt for Specialized Knowledge Collection}
\label{prompt_know}

\noindent \textbf{Generate Learning Summary} only prompts LLMs to analyze the SLM's errors and generate the specialized knowledge of learning summary.

\noindent \textbf{Generate Learning Summary and Supplementary Knowledge} prompts LLMs to analyze the SLM's errors and generate the specialized knowledge of learning summary and Supplementary Knowledge, providing additional relevant background knowledge to further assist SLMs in solving similar complex reasoning tasks in the future.

\newpage

~

\newpage
~~~ \newpage
~~~ \newpage
~~~
\newpage
~~~
\newpage
\begin{tcolorbox}[width=\textwidth, title = {Generate Learning Summary}]

As an excellent educational teacher, your goal is to help students enhance their question-solving abilities.

Based on an understanding and explanation of the question, along with relevant background knowledge, fundamental concepts, and empirical conclusions, please generate a learning summary in a numbered list format that will help students complete the same task in the future.

    ~ 
    
\#\#\# Requirements:

1. Learning summary should outline the thought processes and precautions for addressing student mistakes, including, but not limited to, question comprehension, thought steps and mathematical calculations. It should also provide a summative experience to help students solve similar questions in the future.

2. Ensure that the content is understandable and usable by students, while also being concise and effective.

3. The obtained learning summary should be general and generalized, not aimed at specific questions.

4. Keep these requirements in mind while generating the learning summary and supplementary knowledge.
  
    ~ 
    
\#\#\# Return Format:

Return in the following format:

Learning Summary: [Learning Summary]
  
    ~ 
    
Question: \{QUESTION\}

Answer: \{ANSWER\}

Please follow the requirements and provide the learning summary.
\end{tcolorbox}

\newpage
\newpage
\newpage
\newpage

    ~ 
    
    ~ 
    
    ~ 
    
    ~ 
        ~ 
    
    ~ 
    
    ~ 
    
    ~ 
        ~ 
    
    ~ 
    
    ~ 
      ~ 
    
    ~ 
    
    ~ 
    
    ~ 
        ~ 
    
    ~ 
    
    ~ 
    
    ~ 
        ~ 
    
    ~ 
    
    ~ 
      ~ 
    
    ~ 
    
    ~ 
    
    ~ 
        ~ 
    
    ~ 
    
    ~ 
    
    ~ 
        ~ 
    
    ~ 
    
    ~ 
      ~ 
    
    ~ 
    
    ~ 
    
    ~ 
        ~ 
    
    ~ 
    
    ~ 
    
    ~ 
        ~ 
    
    ~ 
    
    ~ 
 
\begin{tcolorbox}[width=\textwidth, title = {Generate Learning Summary and Supplementary Knowledge}]
As an excellent educational teacher, your goal is to help students enhance their question-solving abilities and to aid students in completing the same task in the future.  
    
You should generate targeted, detailed thought processes and relevant background knowledge for solving similar questions in the future.

Your role involves creating learning summaries and supplementary knowledge, specifically identifying the steps needed to solve the question and providing additional general knowledge in the supplementary knowledge.

    ~ 
    
\#\#\# Requirements:

1. Learning summary should outline the thought processes including, but is not limited to, question comprehension, thought steps and mathematical calculations. It should also provide a summative experience to help students solve similar questions in the future.

2. Supplementary knowledge should include a list of essential background information that students need to solve the question. This should encompass, but is not limited to, mathematical formulas, definitions, relevant world knowledge, and specific techniques.

3. Ensure that the content is understandable and usable by students, while also being concise and effective.

4. The obtained learning summary should be general and generalized, not aimed at specific problems, and the supplementary knowledge should also be general knowledge of the problem without involving specific analysis.

5. Keep these requirements in mind while generating the learning summary and supplementary knowledge.
  
    ~ 
    
\#\#\# Return Format:

Return in the following format:

Learning Summary: [Learning Summary]

Supplementary Knowledge: [Supplementary Knowledge]
  
    ~ 
    
Question: \{QUESTION\}

Answer: \{ANSWER\}

Please follow the requirements and provide the learning summary and supplementary knowledge.
\label{full}
\end{tcolorbox}

\end{document}